\documentclass{article}

    \PassOptionsToPackage{numbers, compress}{natbib}
\usepackage[final]{neurips_2020}

\usepackage[utf8]{inputenc} % allow utf-8 input
\usepackage[T1]{fontenc}    % use 8-bit T1 fonts
\usepackage{hyperref}       % hyperlinks
\usepackage{url}            % simple URL typesetting
\usepackage{booktabs}       % professional-quality tables
\usepackage{amsfonts}       % blackboard math symbols
\usepackage{nicefrac}       % compact symbols for 1/2, etc.
\usepackage{microtype}      % microtypography
\usepackage{enumitem}
\usepackage{graphicx}
\usepackage{tabularx}
\usepackage{adjustbox}
\usepackage{threeparttable}
\usepackage{subfig}
\usepackage{amsmath}
\usepackage{algorithm}
\usepackage{algorithmic}
\usepackage{multirow}
\usepackage{tabu}
\usepackage{amssymb}
\usepackage{varwidth}
\usepackage{wrapfig}
\usepackage{booktabs}

\usepackage{xcolor}
\usepackage{transparent}
\usepackage{tabulary}
\usepackage{amsfonts}
\usepackage{lipsum}
\usepackage{capt-of}% or \usepackage{caption}
\usepackage{wrapfig}

\DeclareMathOperator{\E}{\mathbb{E}}
\DeclareCaptionLabelFormat{singleparen}{#2}
\newcommand{\figref}[1]{Fig.~\ref{#1}}
\newcommand{\tabref}[1]{Table~\ref{#1}}

\newcommand{\secref}[1]{Sec.~\ref{#1}}
\newcommand{\ie}{\textit{i.e.}}
\newcommand{\eg}{\textit{e.g.}}
\newcommand{\etal}{\textit{et.al.}}

\hypersetup{
     colorlinks   = true,
     citecolor    = green
}
\hypersetup{linkcolor=red}

\title{Discover, Hallucinate, and Adapt: Open Compound Domain Adaptation for Semantic Segmentation}

\author{%
  Kwanyong Park,   Sanghyun Woo,   Inkyu Shin,   In So Kweon  \\
  Korea Advanced Institute of Science and Technology (KAIST)\\
  \texttt{$\{$pkyong7,shwoo93,dlsrbgg33,iskweon77$\}$@kaist.ac.kr} \\
}

\begin{document}

\maketitle

\begin{abstract}

Unsupervised domain adaptation (UDA) for semantic segmentation has been attracting attention recently, as it could be beneficial for various label-scarce real-world scenarios (e.g., robot control, autonomous driving, medical imaging, etc.). Despite the significant progress in this field, current works mainly focus on a single-source single-target setting, which cannot handle more practical settings of multiple targets or even unseen targets. 
In this paper, we investigate open compound domain adaptation (OCDA), which deals with mixed and novel situations at the same time, for semantic segmentation.
We present a novel framework based on three main design principles: \textit{discover}, \textit{hallucinate}, and \textit{adapt}. The scheme first clusters compound target data based on style, discovering multiple latent domains (\textbf{discover}). Then, it hallucinates multiple latent target domains in source by using image-translation (\textbf{hallucinate}). This step ensures the latent domains in the source and the target to be paired. Finally, target-to-source alignment is learned separately between domains (\textbf{adapt}). In high-level, our solution replaces a hard OCDA problem with much easier multiple UDA problems.
We evaluate our solution on standard benchmark GTA5 to C-driving, and achieved new state-of-the-art results.
\end{abstract}

\section{Introduction}

Deep learning-based approaches have achieved great success in the semantic segmentation~\cite{long2015fully,zhao2017pyramid,chen2017deeplab,dai2017deformable,zhang2018context,chen2019graph,kim2020video, fu2019dual}, thanks to a large amount of fully annotated data. However, collecting large-scale accurate pixel-level annotations can be extremely time and cost consuming~\cite{cordts2016cityscapes}. An appealing alternative is to use off-the-shelf simulators to render synthetic data for which ground-truth annotations are generated automatically~\cite{richter2016playing,RosCVPR16,richter2017playing}. Unfortunately, models trained purely on simulated data often fail to generalize to the real world due to the \textit{domain shifts}. Therefore, a number of unsupervised domain adaptation (UDA) techniques~\cite{ganin2016domain,tzeng2017adversarial,bousmalis2017unsupervised} that can seamlessly transfer knowledge learned from the label-rich source domain (\textit{simulation}) to an unlabeled new target domain (\textit{real}) have been presented.

%Deep learning for semantic segmentation has shown great progress in recent years with many applications such as autonomous driving and abc. But most methods require a large amount of pixel level label, which is hard to gather, both in time and cost. To address these data issues, with the use of a various synthetic datasets, many unsupervised learning methods ~\cite{adaptseg,advent,cbst,cycada,taking,choi2019self} have been proposed under the name of unsupervised domain adaptation.

Despite the tremendous progress of UDA techniques, we see that their experimental settings are still far from the real-world.
In particular, existing UDA techniques mostly focus on a single-source single-target setting~\cite{adaptseg,advent,cbst,cycada,taking,park2019preserving,choi2019self,pan2020unsupervised}. They do not consider a more practical scenario where the target consists of multiple data distributions without clear distinctions. To investigate a continuous and more realistic setting for domain adaptation, we study the problem of open compound domain adaptation (OCDA)~\cite{compound}. In this setting, the target is a union of multiple homogeneous domains without domain labels. The unseen target data also needs to be considered at the test time, reflecting the realistic data collection from both mixed and novel situations.

\begin{figure*}[t]
\centering
\includegraphics[width=1\textwidth]{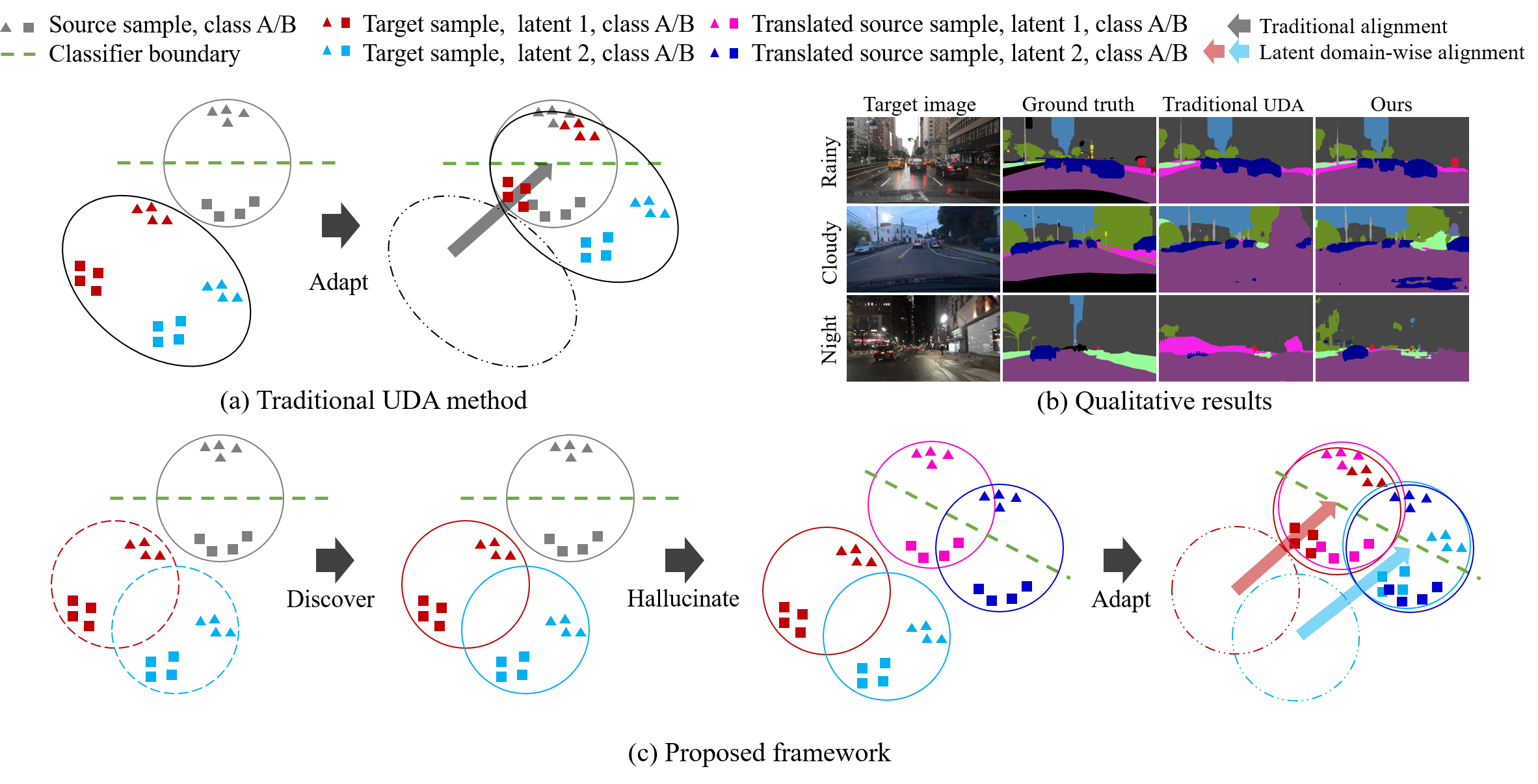}
\captionsetup{font=footnotesize}
\caption{
\textbf{Overview of the proposed OCDA framework: Discover, Hallucinate, and Adapt.} 
The traditional UDA methods consider compound target data as a uni-modal distribution and adapt it at once. 
Therefore, only the target data that is close to the source tends to align well (\textit{biased alignment}). 
On the other hand, the proposed scheme explicitly finds multiple latent target domains and adopts domain-wise adversaries. 
The qualitative results demonstrates that our solution indeed resolves the biased-alignment issues successfully.
We adopt AdaptSeg~\cite{adaptseg} as the baseline UDA method.
}
\label{fig:intro}
 \vspace{-3mm}
\end{figure*}

%Despite the great amount of development and finding that has been made on the unsupervised domain adaptation, the experimental settings used in many works are still far from reality. Most papers treated source and target domain as uni-modal distribution. To narrow the gap between the experimental settings and reality, more realistic UDA settings were proposed from the perspective of source or target domain as complex distributions. multi-source ~\cite{zhao2018adversarial, zhao2019multi}, multi-target ...
%Recently, et.al propose and investigate new settings, named open compound domain adaptation(OCDA). In OCDA problem, target domain can be regarded as a combination of multiple uni-modal distributions without explicit domain labels. In addition, samples may come from novel domain when testing a model.

A naive way to perform OCDA is to apply the current UDA methods directly, viewing the compound target as a uni-modal distribution.
As expected, this method has a fundamental limitation;
It induces a \textit{biased alignment}\footnote{We provide quantitative analysis in \secref{sec:bias_align}.}, where only the target data that are close to source aligns well (see~\figref{fig:intro} and ~\tabref{table:abl_component}-(b)).
We note that the compound target includes various domains that are both close to and far from the source. Therefore, alignment issues occur if multiple domains and their differences in target are not appropriately handled.
Recently, Liu~\etal~\cite{compound} proposed a strong OCDA baseline for semantic segmentation.
The method is based on easy-to-hard curriculum learning~\cite{cbst}, where the easy target samples that are close to the source are first considered, and hard samples that are far from the source are gradually covered.
While the method shows better performance than the previous UDA methods, we see there are considerable room for improvement as they do not fully utilize the domain-specific information\footnote{The OCDA formulation in~\cite{compound} exploits domain-specific information. Though, it is only for the classification task, and the authors instead use a degenerated model for the semantic segmentation task as they cannot access the domain encoder. Please refer to the original paper for the details. This shows that extension of the framework from classification to segmentation (i.e., structured output) is non-trivial and requires significant domain knowledge.}.

% A straightforward way to perform OCDA is to apply a traditional UDA method with the assumption that all target samples from uni-modal distribution. However, due to the complexity and diversity of the compound target domain, the naive approach shows several limitations. For example, night images, which is from a distribution distant from the source domain, show lower performance as training progresses in GTA5 to C-Driving OCDA problems. Liu~\etal ~\cite{compound} propose curriculum learning based on samples’ individual gaps to source domain for handling the mixed distribution of compound domain adaptation. While the proposed curriculum learning shows impressive improvement, estimating the individual gap is not accessible in semantic segmentation.

To this end, we propose a new OCDA framework for semantic segmentation that incorporates three key functionalities: discover, hallucinate, and adapt.
We illustrate the proposed algorithm in~\figref{fig:intro}.
Our key idea is simple and intuitive: decompose a hard OCDA problem into multiple easy UDA problems. 
We can then ease the optimization difficulties of OCDA and also benefit from the various well-developed UDA techniques.
In particular, the scheme starts by discovering $K$ latent domains in the compound target data~\cite{matsuura2019domain} (\textbf{discover}).
Motivated by the previous works~\cite{huang2018multimodal,lee2018diverse,ma2018exemplar,huang2017arbitrary,cho2019image, shen2019towards} that utilizes style information as domain-specific representation, we propose to use \textit{latent target styles} to cluster the compound target.
Then, the scheme generates $K$ target-like source domains by adopting an examplar-guided image translation network~\cite{choi2019self,wang2019example}, hallucinating multiple latent target domains in source (\textbf{hallucinate}).
Finally, the scheme matches the latent domains of source and target, and by using $K$ different discriminators, the domain-invariance is captured separately between domains~\cite{adaptseg,advent} (\textbf{adapt}).
%We empirically observe the proposed training scheme   the successfully resolves the biased-alignment issue of current UDA techniques.

We evaluate our framework on standard benchmark, GTA5~\cite{richter2016playing} to C-driving, and achieved new state-of-the-art OCDA performances.
To empirically verify the efficacy of our proposals, we conduct extensive ablation studies.
We confirm that three proposed design principles are complementary to each other in constructing an accurate OCDA model.

\begin{figure*}[h]
\centering
\includegraphics[width=1\textwidth]{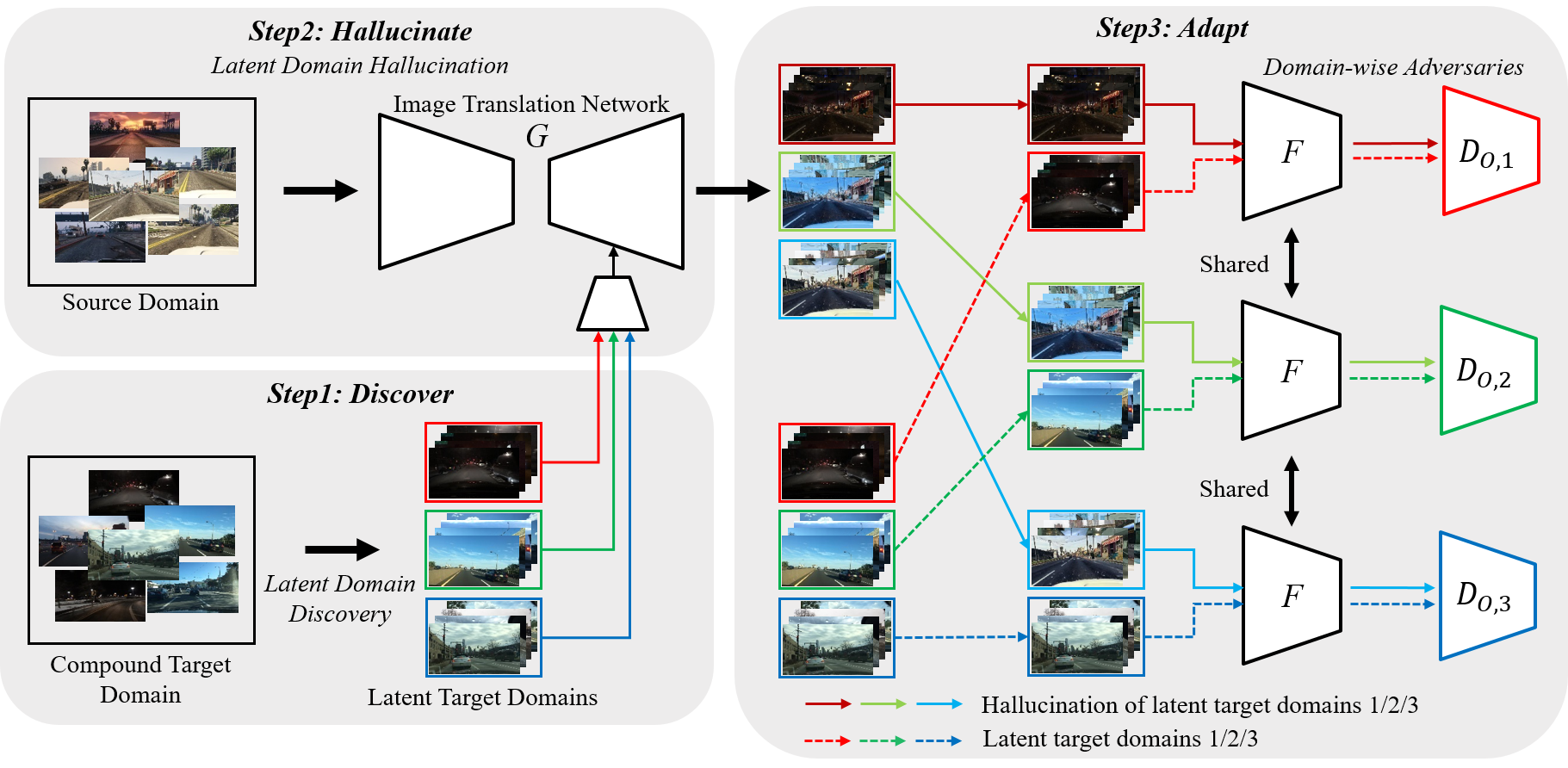}
\captionsetup{font=footnotesize}
\caption{\textbf{Overview of the proposed network.} 
Following the proposed DHA (\textbf{Discover}, \textbf{Hallucinate}, and \textbf{Adapt}) training scheme, the network is composed of three main blocks.
1) \textbf{Discover}: Regarding the `\textit{style}' as domain-specific representation, the network partitions the compound target data into a total of $K$ clusters.
We see each cluster as a specific latent domain. 
2) \textbf{Hallucinate}: In the source domain, the network hallucinates $K$ latent targets using image-translation method.
The source images are then closely aligned with the target, reducing the domain gap in a pixel-level.
3) \textbf{Adapt}: The network utilizes $K$ different discriminators to enforce domain-wise adversaries. 
In this way, we are able to explicitly leverage the latent multi-mode structure of the data.
Connecting all together, the proposed network successfully learns domain-invariance from the compound target.
}
\label{fig:network_overview}
\end{figure*}

\section{Method}

In this work, we explore OCDA for semantic segmentation.
The goal of OCDA is to transfer knowledge from the labeled source domain $S$ to the unlabeled compound target domain $T$, so that trained model can perform the task well on both $S$ and $T$. Also, at the inference stage, OCDA tests the model in open domains that have been previously unseen during training.

% Source domain $S$, $X_{S}=\left \{ \mathbf{x}_{S}^{i} \right \}_{i=1}^{N_{S}}$ , $Y_{S}=\left \{ \mathbf{y}_{S}^{i} \right \}_{i=1}^{N_{S}}$
% Compound target domain $T$, $X_{T}=\left \{ \mathbf{x}_{T}^{i} \right \}_{i=1}^{N_{T}}$ 

\subsection{Problem setup}

We denote the source data and corresponding labels as $X_{S}=\left \{ \mathbf{x}_{S}^{i} \right \}_{i=1}^{N_{S}}$ and $Y_{S}=\left \{ \mathbf{y}_{S}^{i} \right \}_{i=1}^{N_{S}}$, respectively. $N_{S}$ is the number of samples in the source data.
We denote the compound target data as $X_{T}=\left \{ \mathbf{x}_{T}^{i} \right \}_{i=1}^{N_{T}}$, which are from the mixture of multiple homogeneous data distributions. $N_{T}$ is the number of samples in the compound target data. We assume that all the domains share the same space of classes (i.e., closed label set).

%We investigate the problem of open compound domain adaptation. The goal of open compound domain adaptation is to transfer knowledge from the labeled source domain $S$ to the unlabeled compound target domain $T$, so that trained model can perform task well on both $S$ and $T$. We denote the source data and corresponding labels as $X_{S}=\left \{ \mathbf{x}_{S}^{i} \right \}_{i=1}^{N_{S}}$ and $Y_{S}=\left \{ \mathbf{y}_{S}^{i} \right \}_{i=1}^{N_{S}}$ respectively. In compound target domain $T$, the target data $X_{T}=\left \{ \mathbf{x}_{T}^{i} \right \}_{i=1}^{N_{T}}$ from the mixture of multiple homogeneous distribution is only given without task labels.

\subsection{DHA: Discover, Hallucinate, and Adapt}

The overview of the proposed network is shown in~\figref{fig:network_overview}, which consists of three steps: Discover, Hallucinate, and Adapt.
The network first discovers multiple latent domains based on style-based clustering in the compound target data (\textbf{Discover}).
Then, it hallucinates found latent target domains in source by translating the source data (\textbf{Hallucinate}). 
%We then can pair the latent domains in source and target.
%The classifier boundary also changes, since we train the segmentation model on the translated source data. 
Finally, domain-wise target-to-source alignment is learned (\textbf{adapt}).
We detail each step in the following sections.

%The key idea to handle complexity of compound target domain and learning process, we adopt a divide-and-conquer strategy, designed to formulate into a setting similar to multiple traditional (single source to single target) UDA problems with reduced domain gap.
%\park{OR}The key idea to handle multimode of compound target domain and resulting domain gap, we designed to construct mode-wise alignment between target latent mode and it’s imitated counterpart which have much reduced gap to the mode.

%To this end, we propose a novel framework, which is consist of three main steps: \textbf{discover}, \textbf{imitate}, \textbf{adapt}. The overview of the proposed framework is shown in \figref{fig:framework_overview}. 

\subsubsection{Discover: Multiple Latent Target Domains Discovery}

%Conventional multi-target setting~\cite{} assumes there are multiple \textit{explicit} target data distributions with clear distinctions between them (i.e., countable set).
%On the other hand, our compound target is composed of multiple homogeneous data distributions that are blended \textit{implicitly} (i.e., uncountable set).

The key motivation of the discovery step is to make \textit{implicit} multiple target domains \textit{explicit} (see~\figref{fig:intro} (c) - Discover).
To do so, we collect domain-specific representations of each target image and assign pseudo domain labels by clustering (\ie, $k$-means clustering~\cite{kanungo2002efficient}).
In this work, we assume that the latent domain of images is reflected in their \textbf{\textit{style}}~\cite{huang2018multimodal,lee2018diverse,ma2018exemplar,huang2017arbitrary,cho2019image, shen2019towards}, and we thus use style information to cluster the compound target domain.
In practice, we introduce hyperparameter $K$ and divide the compound target domain $T$ into a total of $K$ latent domains by style, $\left \{T_{j}\right \}_{j=1}^{K}$. 
Here, the style information is the convolutional feature statistics (i.e., mean and standard deviations), following~\cite{huang2017arbitrary,dumoulin2016learned}.
After the discovery step, the compound target data $X_{T}$ is divided into a total of $K$ mutually exclusive sets. 
The target data in the $j$-th latent domain ($j \in {1,...,K}$), for example, can be expressed as following: $X_{T,j} =\left \{ \mathbf{x}_{T,j}^{i} \right \}_{i=1}^{N_{T,j}}$, where $N_{T,j}$ is the number of target data in the $j$-th latent domain~\footnote{$X_{T,j}$ and $N_{T,j}$ satisfy $X_{T} = \bigcup_{j=1}^{K}X_{T,j}$ and $\sum_{j}N_{T,j}=N_{T}$, respectively.}.

%Note, the divided target domains are mutually exclusive ($X_{T,l}\cap X_{T,m} = \emptyset$, where $l\neq m$), and thus, $\sum_{j}N_{T,j}=N_{T}$  and $X_{T} = \bigcup_{j=1}^{K}X_{T,j}$.

%In latent domain discovery step (\textbf{discover}), we model the compound target domain as a mixture of multiple latent domains.
%Following ~\cite{huang2018multimodal}, we assume that an image consist of a content component which is shared across latent domains, and a style component specific to each latent domain. From this assumption, we divide compound target domain into $K$ latent domains $\left \{T_{j}\right \}_{j=1}^{K}$ by clustering the style component of each image. Specifically, feature statistics (\ie  mean and standard deviation across the spatial dimension) from convolutional neural network which is pre-trained on Imagenet dataset is utilized as style component.
%Divided $j$th latent domains $X_{T,j}$ is composed of $N_{T,j}$ target samples $\left \{ \mathbf{x}_{T,j}^{i} \right \}_{i=1}^{N_{T,j}}$ and pseudo domain label for $\mathbf{x}_{T,j}^{i}$ is assigned to $j$.

%$K$ latent domains: $\left \{T_{j}\right \}_{j=1}^{K}$,
%$j$th latent domains $X_{T,j} =\left \{ \mathbf{x}_{T,j}^{i} \right \}_{i=1}^{N_{T,j}}$ 
%\park{Add?}

%no intersection $X_{T,l}\cap X_{T,m} = \emptyset$ where $l\neq m$
%union of latent domains = compound target domain $\sum_{j}N_{T,j}=N_{T}$  or $X_{T} = \bigcup_{j=1}^{K}X_{T,j}$

% We divide compound target domain into K latent domains via clustering in terms of style features ~\cite{matsuura2019domain}.

\subsubsection{Hallucinate: Latent Target Domains Hallucination in Source}

We now hallucinate $K$ latent target domains in the source domain.
In this work, we formulate it as image-translation~\cite{liu2017unsupervised, zhu2017unpaired, huang2018multimodal,lee2018diverse}.
For example, the hallucination of the $j$-th latent target domain can be expressed as,
%\begin{equation}
%\begin{split}
    $G(\mathbf{x}_{S}^{i}, \mathbf{x}_{T,j}^{z}) \mapsto \mathbf{x}_{S,j}^{i}.$
%\end{split}
%\label{eqn:hallucinate}
%\end{equation}
Where $\mathbf{x}_{S}^{i} \in X_{S}$, $\mathbf{x}_{T,j}^{z} \in X_{T,j}$, and $\mathbf{x}_{S,j}^{i} \in X_{S,j}$~\footnote{
$X_{S,j} = \left \{ \mathbf{x}_{S,j}^{i} \right \}_{i=1}^{N_{S}}$.} are original source data, randomly chosen target data in $j$-th latent domain, and source data translated to $j$-th latent domain. $G(\cdot)$ is exemplar-guided image-translation network. $z \in 1,...,N_{T,j}$ indicates random index. We note that random selection of latent target data improves model robustness on (target) data scarcity.

% high-resolution images
% source content preservation
% target style reflection

Now, the question is how to design an effective image-translation network, $G(\cdot)$, which can satisfy all the following conditions at the same time.
1) high-resolution image translation, 2) source-content preservation, and 3) target-style reflection.
In practice, we adopt a recently proposed exemplar-guided image-translation framework called TGCF-DA~\cite{choi2019self} as a baseline.
We see it meets two former requirements nicely, as the framework is cycle-free~\footnote{Most existing GAN-based~\cite{goodfellow2014generative} image translation methods heavily rely on cycle-consistency~\cite{zhu2017unpaired} constraint. As cycle-consistency, by construction, requires redundant modules such as a target-to-source generator, they are memory-inefficient, limiting the applicability of high-resolution image translation.} and uses a strong semantic constraint loss~\cite{cycada}.
In TGCF-DA framework, the generator is optimized by two objective functions: $L_{GAN}$, and $L_{sem}$.
We leave the details to the appendicies as they are not our novelty.

Despite their successful applications in UDA, we empirically observe that the TGCF-DA method cannot be directly extended to the OCDA.
The most prominent limitation is that the method fails to reflect diverse target-styles (from multiple latent domains) to the output image and rather falls into mode collapse.
We see this is because the synthesized outputs are not guaranteed to be style-consistent (i.e., the framework lacks style reflection constraints).
To fill in the missing pieces, we present a \textit{style consistency loss}, using discriminator $D_{Sty}$ associated with a pair of target images - either both from same latent domain or not:

% \begin{equation}
% \begin{split}
% L_{Style}^{j}(G,D_{Sty})= \E_{\mathbf{x}_{T,j} \sim X_{T,j}}\left [ logD_{Sty}(\mathbf{x}_{T,j}^{1},\mathbf{x}_{T,j}^{2}) \right ] \\
% + \sum_{l\neq j}\E_{\mathbf{x}_{T,j} \sim X_{T,j},\mathbf{x}_{T,l} \sim X_{T,l}}\left [log(1-D_{Sty}(\mathbf{x}_{T,j},\mathbf{x}_{T,l})) \right ] \\
% + \E_{\mathbf{x}_{S} \sim X_{S}, \mathbf{x}_{T,j} \sim X_{T,j}}\left [log(1-D_{Sty}(\mathbf{x}_{T,j},G(\mathbf{x}_{S},\mathbf{x}_{T,j}))) \right ]
% \end{split}
% \label{eqn:style_gan_loss}
% \end{equation}

\begin{equation}
\begin{split}
L_{Style}^{j}(G,D_{Sty})= \E_{\mathbf{x}_{T,j}^{'} \sim X_{T,j}, \mathbf{x}_{T,j}^{''} \sim X_{T,j}}\left[logD_{Sty}(\mathbf{x}_{T,j}^{'},\mathbf{x}_{T,j}^{''}) \right ] \\
+ \sum_{l\neq j}\E_{\mathbf{x}_{T,j} \sim X_{T,j},\mathbf{x}_{T,l} \sim X_{T,l}}\left [log(1-D_{Sty}(\mathbf{x}_{T,j},\mathbf{x}_{T,l})) \right ] \\
+ \E_{\mathbf{x}_{S} \sim X_{S}, \mathbf{x}_{T,j} \sim X_{T,j}}\left [log(1-D_{Sty}(\mathbf{x}_{T,j},G(\mathbf{x}_{S},\mathbf{x}_{T,j}))) \right ]
\end{split}
\label{eqn:style_gan_loss}
\end{equation}

where $\mathbf{x}_{T,j}^{'}$ and $\mathbf{x}_{T,j}^{''}$ are a pair of sampled target images from same latent domain $j$ (i.e., same style), $\mathbf{x}_{T,j}$, and $\mathbf{x}_{T,l}$ are a pair of sampled target images from different latent domain (i.e., different styles). 
The discriminator $D_{Sty}$ learns awareness of style consistency between pair of images. Simultaneously, the generator G tries to fool $D_{Sty}$ by synthesizing images with the same style to exemplar, $\mathbf{x}_{T,j}$. With the proposed adversarial style consistency loss, we empirically verify that the target style-reflection is strongly enforced.

By using image-translation, the hallucination step reduces the domain gap between the source and the target in a pixel-level. Those translated source images are closely aligned with the compound target images, easing the optimization difficulties of OCDA.
Moreover, various latent data distributions can be covered by the segmentation model, as the translated source data which changes the classifier boundary is used for training (see~\figref{fig:intro} (c) - Hallucinate).

\subsubsection{Adapt: Domain-wise Adversaries}

Finally, given $K$ target latent domains $\left \{T_{j}\right \}_{j=1}^{K}$ and translated $K$ source domains $\left \{S_{j}\right \}_{j=1}^{K}$, the model attempts to learn domain-invariant features.
Under the assumption of translated source and latent targets are both a uni-modal now, one might attempt to apply the existing state-of-the-art UDA methods (\eg  Adaptseg~\cite{adaptseg}, Advent~\cite{advent}) directly.
However, as the latent multi-mode structure is not fully exploited, we see this as sub-optimal and observe its inferior performance.
Therefore, in this paper, we propose to utilize $K$ different discriminators, $D_{O,j}, j \in 1, ..., K$ to achieve (latent) domain-wise adversaries instead.
For example, $j$-th discriminator $D_{O,j}$ only focuses on discriminating the output probability of segmentation model from $j$-th latent domain (i.e., samples either from $T_{j}$ or $S_{j}$).
The adversarial loss for $j$th target domain is defined as:

\begin{equation}
\begin{split}
L_{Out}^{j}(F,D_{O,j})= \E_{\mathbf{x}_{S,j} \sim X_{S,j}}\left [ logD_{O,j}(F(\mathbf{x}_{S,j})) \right ] + \E_{\mathbf{x}_{T,j} \sim X_{T,j}}\left [ log(1-D_{O,j}(F(\mathbf{x}_{T,j}))) \right ]
\end{split}
\label{eqn:DW_discrim_loss}
\end{equation}

where $F$ is segmentation network.
The (segmentation) task loss is defined as standard cross entropy loss.
For example, the source data translated to the $j$-th latent domain can be trained with the original annotation as:

\begin{equation}
\begin{split}
L_{task}^{j}(F)=-\E_{(\mathbf{x}_{S,j},\mathbf{y}_{S})\sim (X_{S,j},Y_{S})}\sum_{h,w}\sum_{c}\mathbf{y}_{s}^{(h,w,c)}log(F(\mathbf{x}_{S,j}))^{(h,w,c)}))
\end{split}
\label{eqn:seg_loss}
\end{equation}

We use the translated source data $\left \{X_{S,j}\right \}_{j=1}^{K}$ and its corresponding labels ${Y}_{s}$.

%one can train a task segmentation model F based on task loss $L_{task}$ with naive adoption of the existing UDA methods (\eg Adaptseg~\cite{adaptseg}, Advent~\cite{advent}), assuming both whole set of adapted source and target latent domains as a uni-modal distribution.

%This naive adoption does not exploiting the multimode structure of domain gap and show inferior performance on some latent domains. Instead we allocate $K$ discriminator,$D_{O,j},j=1,...,K$, for latent domain-wise alignment. 

%$j$-th discriminator $D_{O,j}$ only tries to discriminate between samples from $T_{j}$ and samples from $S_{j}$in output space of $F$. Meanwhile, the task network tries to fool only the $D_{O,j}$ for images from $S_{j}$, $T_{j}$. The output-level adversarial loss for $j$th target domain is defined as:
%\begin{equation}
%\begin{split}
%L_{Out}^{j}(F,D_{O,j})= \E_{\mathbf{x}'_{S,j} \sim X'_{S,j}}\left [ logD_{O,j}(F(\mathbf{x}'_{S,j})) \right ] + \E_{\mathbf{x}_{T,j} \sim X_{T,j}}\left [ log(1-D_{O,j}(F(\mathbf{x}_{T,j}))) \right ]
%\end{split}
%\label{eqn:DW_discrim_loss}
%\end{equation}

% adversarial adaptation $D_{O,j},j=1,...,K$,

% Standard domain adaptation method ~\cite{adaptseg, advent}

%  Segmentation Network M.

% Discriminator training
% \min_{d}

% \begin{equation}
% \begin{split}
% L_{d}(P) = -\sum_{k}1_{[dl=k]}\sum_{h,w}(1-z) \log(D_{k}(P)^{(h,w,0)})+z \log(D_{k}(P)^{(h,w,1)}). 
% \end{split}
% \label{eqn:DW_discrim_lossv2}
% \end{equation}

% \subsection{Adapt: Domain Generalization}

\subsection{Objective Functions}

The proposed DHA learning framework utilizes adaptation techniques, including pixel-level alignment, semantic consistency, style consistency, and output-level alignment.
The overall objective loss function of DHA is:

\begin{equation}
\begin{split}
L_{total} = \sum_{j} \left [ \lambda_{GAN} L_{GAN}^{j} + \lambda_{sem}  L_{sem}^{j} + \lambda_{Style}  L_{Style}^{j} + \lambda_{Out}  L_{Out}^{j} + \lambda_{task}  L_{task}^{j} \right ]
\end{split}
\label{eqn:total_loss}
\end{equation}

Here, we use $\lambda_{GAN}=1$, $\lambda_{sem}=10$, $\lambda_{Style}=10$, $\lambda_{out}=0.01$, $\lambda_{task}=1$.
Finally, the training process corresponds to solving the following optimization,
% \begin{equation}
% \begin{split}
$ F^{*} = \arg \min_{F} \min_{D} \max_{G} L_{total}$,
% \end{split}
% \label{eqn:seg_model}
% \end{equation}
where G and D represents a generator (in $L_{sem}$, $L_{GAN}$, and $L_{Style}$) and all the discriminators (in $L_{GAN}$, $L_{Style}$, and $L_{Out}$), respectively.

\section{Experiments}

In this section, we first introduce experimental settings and then compare the segmentation results of the proposed framework and several state-of-the-art methods both quantitatively and qualitatively, followed by ablation studies.

\subsection{Experimental Settings}

\noindent{\textbf{Datasets.}}
In our adaptation experiments, we take GTA5~\cite{richter2016playing} as the source domain, while the BDD100K dataset~\cite{yu2018bdd100k} is adopted as the compound (``rainy'', ``snowy'', and ``cloudy'') and open domains (``overcast'') (\ie, C-Driving~\cite{compound}).

\noindent{\textbf{Baselines.}}
We compare our framework with the following methods. 
\textbf{(1) Source-only,} train the segmentation model on the source domains and test on the target domain directly.
\textbf{(2) UDA methods,} perform OCDA via (single-source single-target) UDA, including AdaptSeg~\cite{adaptseg}, CBST~\cite{cbst}, IBN-Net~\cite{pan2018two}, and PyCDA~\cite{lian2019constructing}.
\textbf{(3) OCDA method,} Liu \textit{et.al.}~\cite{compound}, which is a recently proposed curriculum-learning based~\cite{cbst} strong OCDA baseline.

\noindent{\textbf{Evaluation Metric.}}
We employ standard mean intersection-over-union (mIoU) to evaluate the segmentation results. We report both results of individual domains of compound(``rainy'', ``snowy'', ``cloudy'') and open domain(``overcast'') and averaged results.

\noindent{\textbf{Implementation Details.}}
\begin{itemize}
    \item \textbf{Backbone}
    We use a pre-trained VGG-16~\cite{simonyan2014very} as backbone network for all the experiments. 
    \item \textbf{Training}
    By design, our framework can be trained in an end-to-end manner.
    However, we empirically observe that splitting the training process into two steps allows stable model training.
    In practice, we cluster the compound target data based on their style statistics (we use ImageNet-pretrained VGG model~\cite{simonyan2014very}).
    With the discovered latent target domains, we first train the hallucination step.
    Then, using both the translated source data and clustered compound target data, we learn the target-to-source adaptation. 
    We adopt two different training schemes (short and long) for the experiments.
    For the short training scheme (5K iteration), we follow the same experimental setup of~\cite{compound}. 
    For the longer training scheme (150K iteration), we use LS GAN~\cite{mao2017least} for Adapt-step training. 
    %with $\lambda_{out}=0.01$.
    %\woo{details on short/long training schemes}
    \item \textbf{Testing}
    We follow the conventional inference setup~\cite{compound}.
    Our method shows superior results against the recent approaches without any overhead in test time.
\end{itemize}

% Theoretically, hallucination and adaptation step could be trained in an end-to-end manner with target latent domains found in discovery step and overall loss function. However, for stable training, we split the training process of hallucination and adaptation step and construct HDA framework in three step. Conceptually, our DHA framework utilize any traditional UDA method~\cite{} for constructing Adapt step. However, we are based on Adaptsegnet~\cite{} for pair comparison with ~\cite{}, if not specified.

\subsection{Comparison with State-of-the art}
%over conventional UDA approaches and recent curriculum-based OCDA method.
%benefiting from fully utilizing of domain specific information in rigorously designed three steps. 
%The results consistently shows that superiority of our method against these models without any overhead in test time.

\begin{figure*}[h]
\centering
\includegraphics[width=\textwidth]{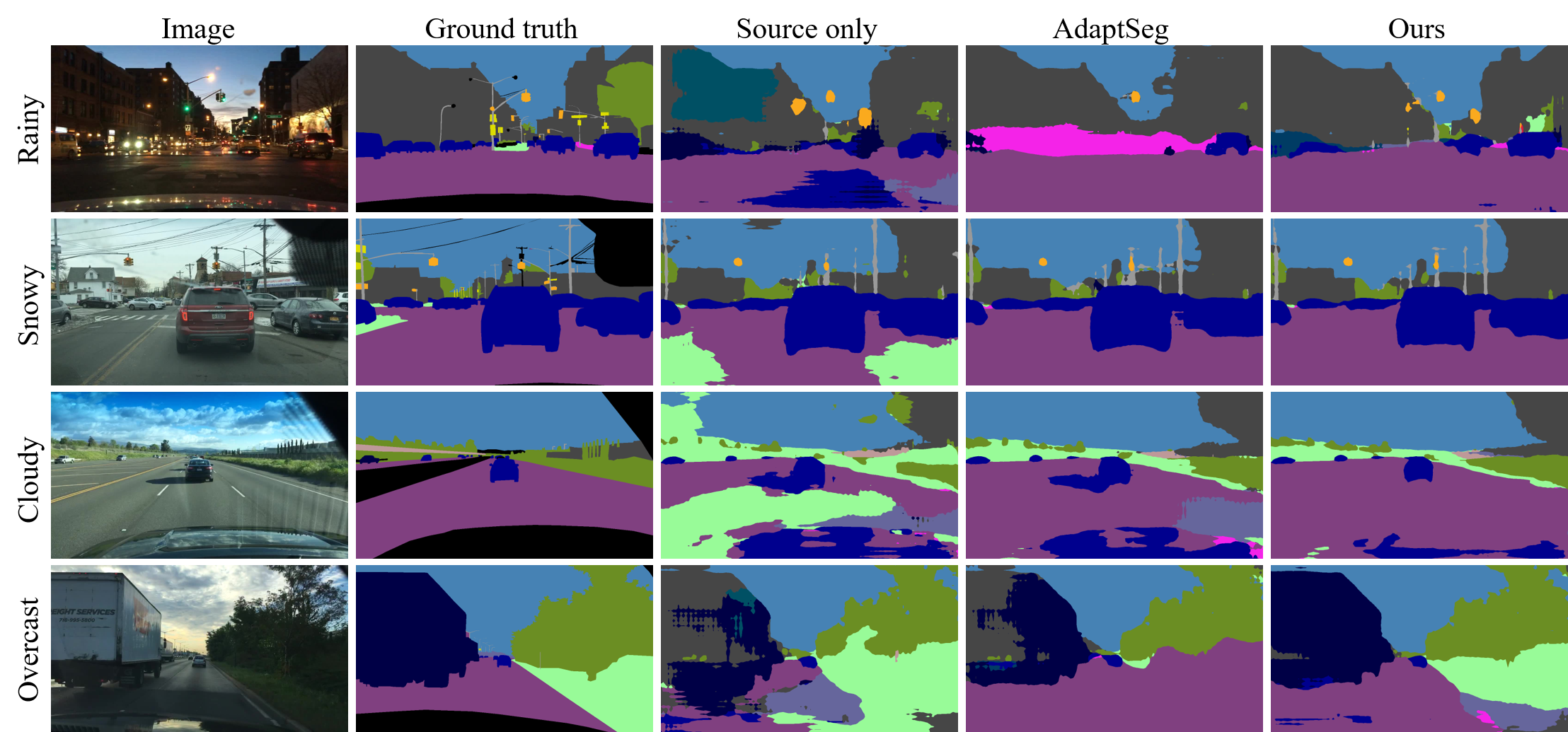}
\caption{\textbf{Qualitative results comparison} of semantic segmentation on the compound domain(``rainy'', ``snowy'', ``cloudy'') and open domain(``overcast''). We can observe clear improvement against both source only and traditional adaptation models~\cite{adaptseg}.
}
\label{fig:seg_qual}
% \vspace{-5mm}
\end{figure*}

We summarize the quantitative results in~\tabref{table:SOTA_comparison}. 
we report adaptation performance on GTA5 to C-Driving.
We compare our method with Source-only model, state-of-the-art UDA-models~\cite{adaptseg,cbst,pan2018two,lian2019constructing,advent}, and recently proposed strong OCDA baseline model~\cite{compound}.
We see that the proposed DHA framework outperforms all the existing competing methods, demonstrating the effectiveness of our proposals.
%The results show that our new solution is indeed general and robust.
We also provide qualitative semantic segmentation results in ~\figref{fig:seg_qual}. 
We can observe clear improvement against both source only and traditional adaptation models~\cite{adaptseg}.

We observe adopting a longer training scheme improves adaptation results ($\dagger$ in~\tabref{table:SOTA_comparison} indicates models trained on a longer training scheme).
Nevertheless, our approach consistently brings further improvement over the baseline of source-only, which confirms its enhanced adaptation capability.
Unless specified, we conduct the following ablation experiments on the longer-training scheme.

\begin{table}[t]
\setlength{\tabcolsep}{1pt}
 \centering
 \caption{\textbf{Comparison with the state-of-the-art UDA/OCDA methods and Ablation study on framework design.} We evaluate the semantic segmentation results, GTA5 to C-driving. (a) $\dagger$ indicates the models trained on a longer training scheme. (b) ``+trad'' denote adopting traditional unsupervised method~\cite{adaptseg}}
    \subfloat[\scriptsize Comparison with the state-of-the-art UDA/OCDA methods]
    {
        \resizebox{0.48\textwidth}{!}
        {
        \def\arraystretch{1.1}
        \begin{tabular}{c|ccc|c|cc}
        \hline
        \multicolumn{1}{c}{Source} & \multicolumn{3}{c}{Compound(C)} & \multicolumn{1}{c}{Open(O)} & \multicolumn{2}{c}{Avg.} \\
        GTA5 & Rainy & Snowy & Cloudy & Overcast & C  & C+O \\
        \hline
        \hline
        Source Only & 16.2 & 18.0 & 20.9 & 21.2 & 18.9 & 19.1  \\
        \hline
        AdaptSeg~\cite{adaptseg} & 20.2 & 21.2 & 23.8 & 25.1 & 22.1 & 22.5  \\
        CBST~\cite{cbst} & 21.3 & 20.6 & 23.9 & 24.7 & 22.2 & 22.6 \\
        IBN-Net~\cite{pan2018two} & 20.6 & 21.9 & 26.1 & 25.5 & 22.8 & 23.5 \\
        PyCDA~\cite{lian2019constructing} & 21.7 & 22.3 & 25.9 & 25.4 & 23.3 & 23.8 \\
        Liu~\etal~\cite{compound} & 22.0 & 22.9 & 27.0 & 27.9 & 24.5 & 25.0 \\
        \hline
        % Source only & 0.0 & 0.0 & 0.0 & 0.0 & 0.0 & 0.0 \\
        Ours & 27.0 & 26.3 & 30.7 & 32.8 & \textbf{28.5} & \textbf{29.2} \\
        \hline
        Source only$\dagger$ & 23.3 & 24.0 & 28.2 & 30.2 & 25.7 & 26.4 \\
        Ours$\dagger$ & 27.1 & 30.4 & 35.5 & 36.1 &\textbf{32.0} & \textbf{32.3} \\
        \hline
        \end{tabular}
        }
  }
      \subfloat[\scriptsize Ablation study on framework design. ]
    { 
            \resizebox{0.50\textwidth}{!}
            {
            \def\arraystretch{1.3}
            \begin{tabular}{c|ccc|cc}
            \hline
            % \multicolumn{1}{c}{Method} & \multicolumn{1}{c}{Discover} & \multicolumn{1}{c}{Hallucinate} & \multicolumn{1}{c}{Adapt} & \multicolumn{1}{c}{C}& & \multicolumn{1}{c}{C+O} \\
            Method & Discover & Hallucinate & Adapt & C & C+O \\
            \hline
            \hline
            Source Only & & &                      & 25.7 & 26.4\\
            % \hline
            Traditional UDA~\cite{adaptseg} & & & +trad                     & 28.8 & 29.3 \\
            % \hline
            (1) &\checkmark &           &  \checkmark                    & 31.1 & 31.1\\
            (2) &\checkmark &\checkmark &                      & 29.8 & 30.4\\
            (3) &\checkmark &\checkmark &+trad       & 30.1 & 31.0\\
            Ours &\checkmark &\checkmark &\checkmark  & \textbf{32.0} & \textbf{32.3}\\
            \hline
            \end{tabular}
            }
  }
\captionsetup{font=footnotesize}
\label{table:SOTA_comparison}
\end{table}

\subsection{Ablation Study}
We run an extensive ablation study to demonstrate the effectiveness of our design choices.
The results are summarized in~\tabref{table:SOTA_comparison}-(b) and ~\tabref{table:abl_component}.
Furthermore, we additionally report the night domain adaptation results (We see the night domain as one of the representative latent domains that are distant from the source).

%While some night images are part of 
%We conduct ablation study in two-level.
%First, we ablate in step-wise and investigate efficacy of each step. This ablation study give us to high level concept of proposed DHA framework.
%Second, we also run detailed ablation study to show the effective design of each step, including effective number of latent target domains in discovery step, effect of style consistency loss in hallucination step, domain-wise adversaries in adapt step.
%In ablation study, we also report the result of night domain which is a representative of target latent domain distant from source. While some night images are part of compound validation set, the effect of it’s performance on final score is negligible, due to the less number of labeled night images. For this reason, we decided to separately report the result of night domain and it give a chance to further analyze the effect of each component. 

\noindent{\textbf{Framework Design.}}
In this experiment, we evaluate three main design principles: \textit{Discover}, \textit{Hallucinate}, and \textit{Adapt}.
We set the adaptation results of both Source Only and Traditional UDA~\cite{adaptseg} as baselines.
First, we investigate the importance of Discover stage (Method (1) in~\tabref{table:SOTA_comparison}-(b)).
The method-(1) learns target-to-source alignment for each clustered latent target domain using multiple discriminators.
As improved results indicate, explicitly clustering the compound data and leveraging the latent domain information allows better adaptation.
Therefore, we empirically confirm our `\textit{cluster-then-adapt}' strategy is effective.
We also explore the Hallucination stage (Method (2) and (3) in~\tabref{table:SOTA_comparison}-(b)).
The method-(2) can be interpreted as a strong Source Only baseline that utilizes translated target-like source data.
The method-(3) further adopts traditional UDA on top of it.
We see both (2) and (3) outperform Source Only and Traditional UDA adaptation results, showing that hallucination step indeed reduces the domain gap.
By replacing the Traditional UDA in method-(3) with the proposed domain-wise adversaries (Ours in~\tabref{table:SOTA_comparison}-(a)), we achieve the best result. The performance improvement of our final model over the baselines is significant.
Note, the final performance drops if any of the proposed stages are missing.
This implies that the proposed three main design principles are indeed complementary to each other.

\begin{figure*}[t]
\centering
\includegraphics[width=1\textwidth]{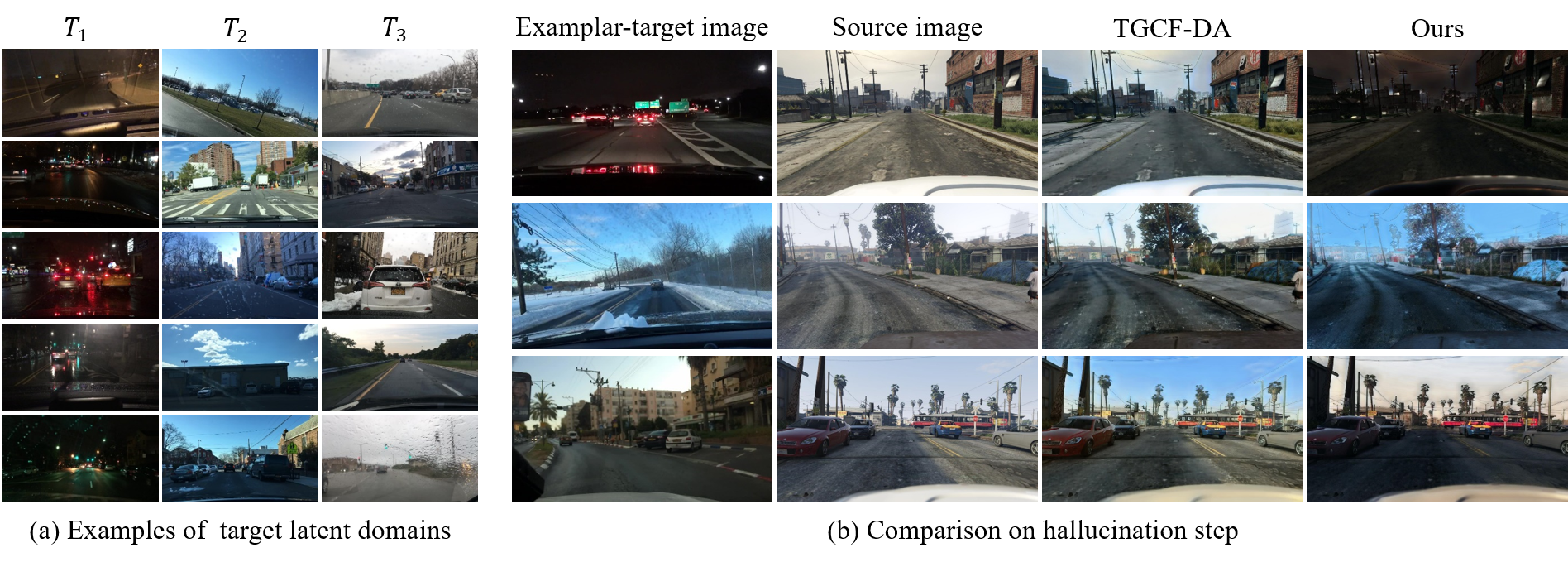}
\caption{\textbf{Examples of target latent domains and qualitative comparison on hallucination step.} (a) We provide random images from each three latent domain (i.e., $K=3$). Note that they have their own `style.' (b) We show the effect of proposed style-consistency loss by comparing ours with original TGCF-DA~\cite{choi2019self} method. }
\label{fig:examplar}
% \vspace{-5mm}
\end{figure*}

\noindent{\textbf{Effective number of latent target domains.}}
%The number of target domains is unknown in the OCDA setting.
%The natural question is then `how many (latent) domains should be found in the given compound target data?'
In this experiment, we study the effect of latent domain numbers ($K$), a hyperparameter in our model.
%We also provide new qualitative insights.
We summarize the ablation results in~\tabref{table:abl_component}-(a).
We vary the number of $K$ from 2 to 5 and report the adaptation results in the Hallucination Step.
As can be shown in the table, we note that all the variants show better performance over the baseline (implying that the model performance is robust to the hyperparameter $K$), and the best adaptation results are achieved with $K=3$.
The qualitative images of found latent domains are shown in~\figref{fig:examplar}-(a).
We can observe that the three discovered latent domains have their own `style.' Interestingly, even these styles (e.g., $T_{1}$: night, $T_{2}$: clean blue,  $T_{3}$: cloudy) do not exactly match the original dataset styles (e.g., ``rainy'', ``snowy'', ``cloudy''), adaptation performance increases significantly.
This indicates there are multiple implicit domains in the compound target by nature, and the key is to find them well and properly handling them.
For the following ablation study, we set $K$ to 3.
%Finding the latent domains in a data-driven manner is an interesting future direction.

%Since the number of latent target domains is unknown in OCDA setting, we have to decide how many latent domains the compound target domain will be divided into. Note that division of compound target domain(~\ie, domain label) is explicitly used to define whether the pair of target images is the same style or not. So, it may be crucial for style consistency learning of $D_{Sty}$ and following steps.

%To verify this issue, we conduct hallucination step with varying the number of latent target domains. With the hallucinated latent domains, for each $K$, we train a semantic segmentation model with ~\eqnref{eqn:seg_loss}(\ie pixel-level adaptation) and report the results in ~\tabref{tab:step2_abl}. 
%With any number of $K$, the performances are superior than the model trained on source dataset. It prove that our hallucination step successfully reduce the domain gap between source and target in pixel-level.
%The result show that $K=3$ is optimal number of latent target domains for C-driving dataset, which shows best performance among various $K$ in terms of averaged result of compound and overcast domain. For following ablation study, we mainly explore with $K=3$.

\noindent{\textbf{Style-consistency loss.}}
If we drop the style consistency loss in the hallucination step, our generator degenerates to the original TGCF-DA~\cite{choi2019self} model.
The superior adaptation results of our method over the TGCF-DA~\cite{choi2019self} in~\tabref{table:abl_component}-(a) implicitly back our claim that the target style reflection is not guaranteed on the original TGCF-DA formulation while ours does.
In~\figref{fig:examplar}-(b), we qualitatively compare the translation results of ours and TGCF-DA~\cite{choi2019self}. 
We can obviously observe that the proposed style-consistency loss indeed allows our model to reflect the correct target styles in the output.
This implies that the proposed solution enforces strong target-style reflection constraints effectively.

%Without the style consistency loss, our hallucination step degenerate into TGCF-DA.  As shown in ~\figref{}, our model successfully reflects style of examplar target image, but TGCF-DA fails to reflects style and fall into mode-collapse. This is because TGCF-DA does not utilize domain information in training of image translation. By injecting style of examplar into synthsized images, our hallucination step further hallucinate distribution of target latent domain, thus trained task model consistently show better performance than trained model on result of TGCF-DA across $K$.

\noindent{\textbf{Domain-wise adversaries.}} 
Finally, we explore the effect of the proposed domain-wise adversaries in~\tabref{table:abl_component}-(b).
We compare our method with the UDA approaches, which consider both the translated source and compound target as uni-modal and thus do not consider the multi-mode nature of the compound target.
While not being sensitive to any specific adaptation methods (i.e., different UDA approaches such as Adaptseg~\cite{adaptseg} or Advent~\cite{advent}), our proposal consistently shows better adaptation results over the UDA approaches.
This implies that leveraging the latent multi-mode structure and conducting adaptation for each mode can ease the complex one-shot adaptation of compound data.

% \begin{table*}[h]
% \caption{\textbf{Ablation Study2: DA, DG.} GTA2Compound, K=3, VGG}
% \begin{center}
% \resizebox{0.90\textwidth}{!}{
% \def\arraystretch{1.1}
% \begin{tabular}{ccc|cccc|c|cc}
% \hline
% \multicolumn{3}{c}{} & \multicolumn{4}{c}{Compound(C)} & \multicolumn{1}{c}{Open(O)} & \multicolumn{2}{c}{Avg.}
% &Source &DA &DG & Rainy & Snowy & Cloudy &Night* & Overcast & C  & C+O \\
% \hline
% \hline
% Ours(K=3) &None &None & 0.0 & 0.0 & 0.0 & 0.0 & 0.0 & 0.0 & 0.0  \\
% \hline
% Ours(K=3) &Adaptsegent &None & 0.0 & 0.0 & 0.0 & 0.0 & 0.0 & 0.0 & 0.0  \\
% Ours(K=3) &Advent &None & 0.0 & 0.0 & 0.0 & 0.0 & 0.0 & 0.0 & 0.0 \\
% \hline
% Ours(K=3) &Ours(Adaptsegent) &None & 0.0 & 0.0 & 0.0 & 0.0 & 0.0 & 0.0 & 0.0  \\
% Ours(K=3) &Ours(Advent) &None & 0.0 & 0.0 & 0.0 & 0.0 & 0.0 & 0.0 & 0.0 \\
% \hline
% Ours &Ours(Adaptsegent) &Ours & 0.0 & 0.0 & 0.0 & 0.0 & 0.0 & 0.0 & 0.0  \\
% Ours &Ours(Advent) &Ours & 0.0 & 0.0 & 0.0 & 0.0 & 0.0 & 0.0 & 0.0 \\
% \hline
% \end{tabular}
% }
% \end{center}
% \label{tab:Abl1}
% \end{table*}

%Model:VGG number of latent domains: 3,4
\begin{table}[t]
\setlength{\tabcolsep}{1pt}
 \centering
 \caption{\textbf{(a)Ablation Study on the Discovery and Hallucination Step.}  We conduct parameter analysis on $K$ to decide the optimal number of latent target domains. Also, we empirically verify the effectiveness of the proposed $L_{Style}$, outperforming TGCF-DA~\cite{choi2019self} significantly. (b)\textbf{Ablation Study on the Adapt step.} We confirm the efficacy of the proposed domain-wise  adaptation, demonstrating its superior adaptation results over the direct application of UDA methods~\cite{adaptseg,advent}.}
    \subfloat[\scriptsize Discovery and Hallucination Step]
         {
         \resizebox{0.50\textwidth}{!}
        {
        \def\arraystretch{1.3}
        \begin{tabular}{c|cccc|c|cc}
        \hline
        \multicolumn{1}{c}{Source} & \multicolumn{4}{c}{Compound(C)} & \multicolumn{1}{c}{Open(O)} & \multicolumn{2}{c}{Avg.} \\
        GTA5 & Rainy & Snowy & Cloudy &Night & Overcast & C  & C+O \\
        \hline
        \hline
        Source Only & 23.3 & 24.0 & 28.2 &8.1 & 30.2 & 25.7 &  26.4\\
        \hline
        TGCF-DA~\cite{choi2019self} & 25.5 & 24.9 & 30.7 & 9.7 & 32.9 & 27.8 & 28.5 \\
        \hline
        Ours(K=2) & 26.0 & 26.6 & 32.4  & 11.1 & 33.6 & 29.3 & 29.7\\
        Ours(K=3) & \textbf{26.4} & \textbf{27.5} & \textbf{33.3}  & 11.8 & \textbf{34.3} & \textbf{29.8} & \textbf{30.4}\\
        Ours(K=4) & 25.2 & 26.4 & 32.7  & 12.1 & 33.8 & 29.1 & 29.5 \\
        Ours(K=5) & 25.4 & 27.0 & 32.5  & \textbf{13.3} & 33.1 & 29.2 & 29.5 \\
        \hline
        \end{tabular}
        }
         
         }
      \subfloat[\scriptsize Adapt Step]
    { 
        \resizebox{0.48\textwidth}{!}
        {
            \def\arraystretch{1.5}
            \begin{tabular}{cc|cccc|c|cc}
            \hline
            \multicolumn{2}{c}{} & \multicolumn{4}{c}{Compound(C)} & \multicolumn{1}{c}{Open(O)} & \multicolumn{2}{c}{Avg.} \\
            Source &Adapt & Rainy & Snowy & Cloudy & Night & Overcast & C  & C+O \\
            \hline
            \hline
                Ours &None & 26.4 & 27.5 & 33.3  & 11.8 & 34.3 & 29.8 & 30.4\\
                \hline
            	Ours &Traditional(~\cite{adaptseg}) & 25.8 & 29.2 & 33.3 & 11.5 & 35.9 & 30.1 &  31.0  \\
            	Ours &Traditional(~\cite{advent}) & 26.7 & 28.9 & 34.7 &12.9& 34.9 & 31.2 & 31.3\\
            	\hline
            	Ours &Domain-wise(~\cite{adaptseg})  & 27.1 & 30.4 & 35.5 & 12.4 & 36.1 &\textbf{32.0} & \textbf{32.3}  \\
                Ours &Domain-wise(~\cite{advent})  & 27.6 & 30.6 & 35.5 & 14.0 & 36.3 & \textbf{32.2} & \textbf{32.5}\\
            % 	\hline
            % 	Ours &DW:AdaptSeg+DG & 0.0 & 0.0 & 0.0 & 0.0 & 0.0 & 0.0  \\
            % 	Ours &DW:Advent+DG & 0.0 & 0.0 & 0.0 & 0.0 & 0.0 & 0.0  \\
            \hline
            \end{tabular}
            \label{table:Abl2_3K}
        }
  }
\captionsetup{font=footnotesize}
\label{table:abl_component}
\end{table}

\subsection{Further Analysis}
 \label{sec:bias_align}
 
\noindent{\textbf{Quantitative Analysis on Biased Alignment.}} In ~\figref{fig:intro}, we conceptually show that the traditional UDA methods induce \textit{biased alignment} on the OCDA setting.
We back this claim by providing quantitative results.
We adopt two strong UDA methods, AdaptSeg~\cite{adaptseg} and Advent~\cite{advent} and compare their performance with ours in GTA5~\cite{richter2016playing} to the C-driving~\cite{compound}.
By categorizing the target data by their attributes, we analyze the adaptation performance in more detail. In particular, we plot the performance/iteration for each attribute group separately. 
% (see~\figref{fig:biased}).

We observe an interesting tendency;
With the UDA methods, the target domains close to the source are well adapted. However, in the meantime, the adaptation performance of distant target domains are compromised~\footnote{We see ``cloudy-daytime'', ``snowy-daytime'', and ``rainy-daytime'' as target domains close to the source, whereas ``dawn'' and ``night'' domain are distant target domains.}. 
In other words, the easy target domains dominate the adaptation, and thus the hard target domains are not adapted well (i.e., biased-alignment).
On the other hand, the proposed DHA framework explicitly discovers multiple latent target domains and uses domain-wise adversaries to resolve the biased-alignment issue effectively. 
We can see that both the close and distant target domains are well considered in the adaptation (i.e., there is no performance drop in the distant target domains).

\begin{wrapfigure}{R}{0.5\textwidth}
\centering
\vspace{-5mm}
\includegraphics[width=0.45\textwidth]{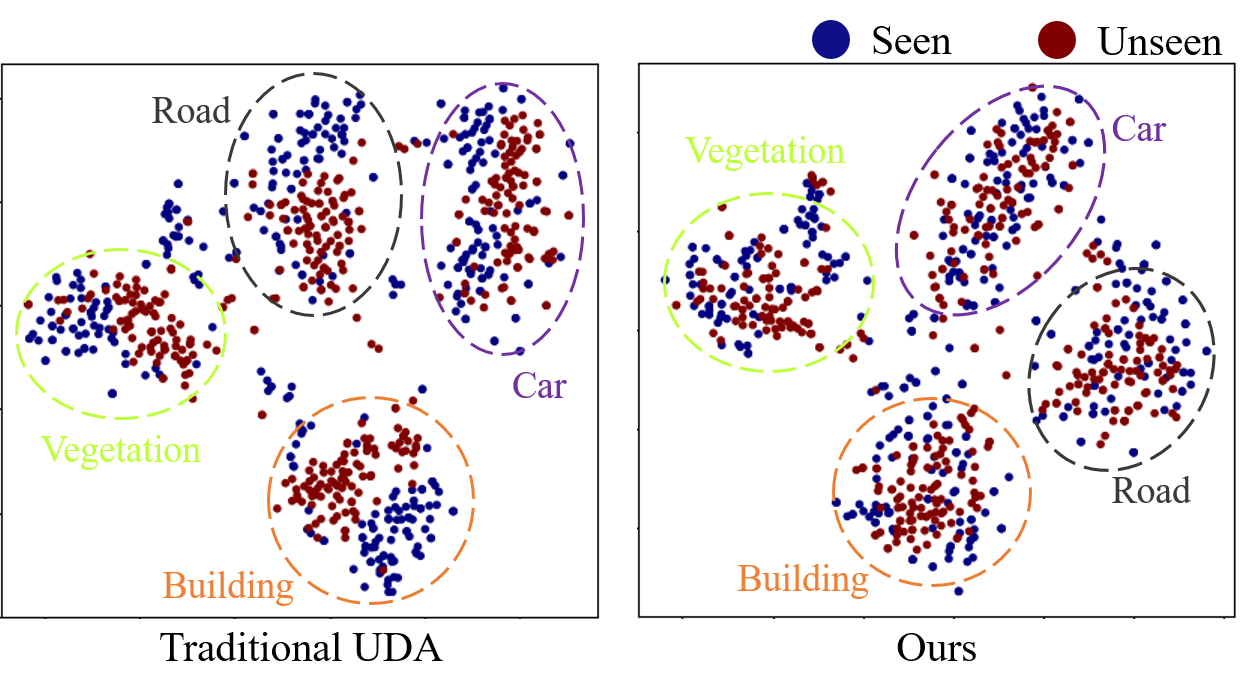}
\vspace{-2mm}
\caption{\label{fig:fig1} t-SNE visualization.}
\vspace{-5mm}
\end{wrapfigure}

\noindent{\textbf{Connection to Domain Generalization.}}
Our framework aims to learn domain-invariant representations that are robust on multiple latent target domains.
As a result, the learned representations can well generalize on the unseen target domains (i.e., open domain) by construction. %(please also refer to the~\figref{fig:fig1}-(b)).
The similar learning protocols can be found in recent domain generalization studies~\cite{li2018domain,matsuura2019domain,dou2019domain,li2018deep} as well.

We analyze the feature space learned with our proposed framework and the traditional UDA baseline~\cite{advent} in the~\figref{fig:fig1}.
It shows that our framework yields more generalized features.
More specifically, the feature distributions of seen and unseen domains are indistinguishable in our framework while not in traditional UDA~\cite{advent}.

\begin{figure*}[t]
\caption{\textbf{Biased-alignment of UDA methods on OCDA.} The following graphs include testing mIoUs of traditional UDA methods~\cite{adaptseg,advent} and ours on GTA5 to C-driving setting. Note that the UDA methods~\cite{adaptseg,advent} tend to induce biased-alignment, where the target domains close to the source are mainly considered for adaptation. As a result, the performance of distant target domains such as ``dawn'' and ``night'' drops significantly as iteration increases. On the other hand, our method resolves this issue and adapts both close and distant target domains effectively.}
\centering
\includegraphics[width=1\textwidth]{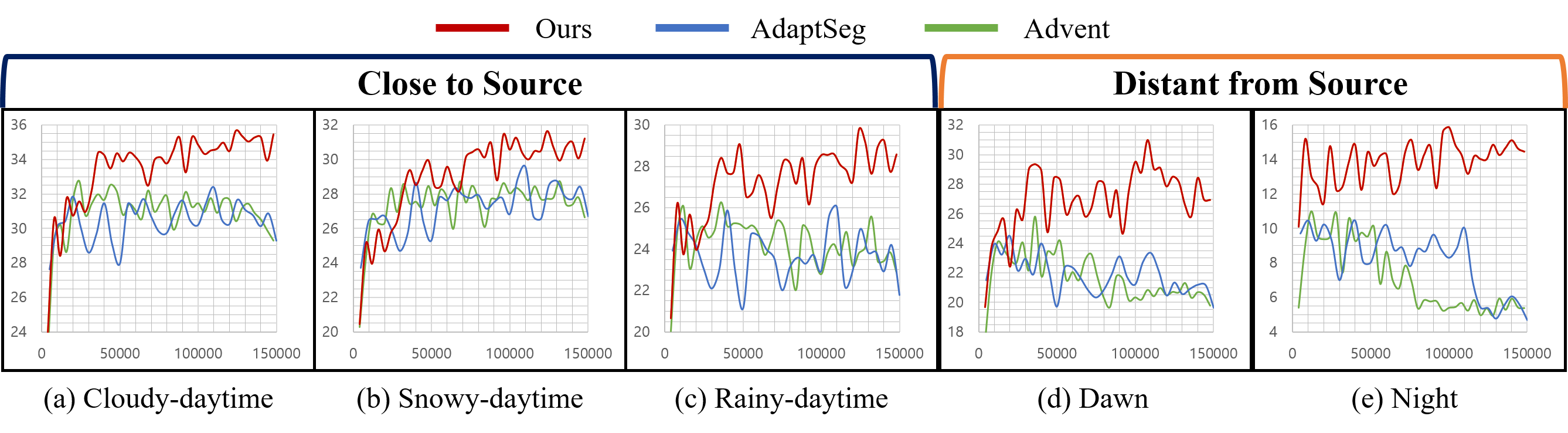}
\label{fig:biased}
% \vspace{-5mm}
\end{figure*}

\section{Conclusion}

In this paper, we present a novel OCDA framework for semantic segmentation.
In particular, we propose three core design principles: Discover, Hallucinate, and Adapt.
First, based on the latent target styles, we cluster the compound target data.
Each group is considered as one specific latent target domain.
Second, we hallucinate these latent target domains in the source domain via image-translation.
The translation step reduces the domain gap between source and target and changes the classifier boundary of the segmentation model to cover various latent domains.
Finally, we learn the target-to-source alignment domain-wise, using multiple discriminators.
Each discriminator focuses only on one latent domain. Finally, we achieve to decompose OCDA problem into easier multiple UDA problems.
Combining all together, we build a strong OCDA model for semantic segmentation.
Empirically, we show that the proposed three design principles are complementary to each other.
Moreover, the framework achieved new state-of-the-art OCDA results, outperforming the existing learning approaches significantly.

\section*{Acknowledgements}
This work was supported by Samsung Electronics Co., Ltd

\section*{Broader Impact}

We investigate the newly presented problem called open compound domain adaptation (OCDA).
The problem well reflects the nature of real-world that the target domain often include mixed and novel situations at the same time.
The prior work on this OCDA setting mainly focuses on the classification task.
Though, we note that extending the classification model to the structured prediction task is non-trivial and requires significant domain-knowledge.
In this work, we identify the challenges of OCDA in semantic segmentation and carefully design a new strong baseline model.
Specifically, we present three core design principles: Discover, Hallucinate, and Adapt.
We empirically show that our proposals are complementary to each other in constructing a strong OCDA model.
We provide both the quantitative and qualitative results to show the efficacy of our final model.
We hope the proposed new algorithm and its results will drive the research directions to step forward towards generalization in the real-world.

{\small
\bibliographystyle{plain}
\bibliography{egbib}
}

\appendix

\section{Appendix}

In this supplementary material, we provide more details about the model and experiments in the following order:
\begin{itemize}
    \item In \secref{sec:other_data}, we evaluate our framework on two new datasets, Synscapes and SYNTHIA, demonstrating that our framework is general.
    % \item In \secref{sec:bias_align}, we provide quantitative results to back our claim that UDA methods induce biased alignment when they are applied to the OCDA setting directly.
    \item In \secref{sec:ablation}, we conduct additional ablation studies on the adaptation step using four latent target domains (i.e., $K=4$). We again see that the proposed domain-wise adversaries outperform the UDA approaches.
    \item In \secref{sec:analysis_K}, we analyze hyperparameter K selection.
    \item In \secref{sec:more_qual}, we show more qualitative results.
    \item In \secref{sec:imple_detail}, we elaborate the implementation details.
\end{itemize}

\subsection{DHA Framework on Other Datasets}
 \label{sec:other_data}
 
We conduct OCDA semantic segmentation experiments using two additional benchmarks: Synscapes~\cite{wrenninge2018synscapes} and SYNTHIA~\cite{RosCVPR16}.
We adopt the source-only method and the state-of-the-art UDA methods~\cite{adaptseg, advent, cbst, zou2019confidence} as baselines.
The adaptation results are summarized in the~\tabref{table:supple_SOTA_comparison}.
We observe that our method consistently outperforms previous UDA approaches on both datasets.
This implies that our DHA framework is indeed general and practical for OCDA.

\subsection{Additional Ablation Study on the Adapt Step}
 \label{sec:ablation}

In the main paper, we already show that the proposed domain-wise adversaries are more effective than the traditional UDA approaches.
To provide more experimental evidence, we conduct an additional ablation study using four latent target domains (i.e., $K=4$).
The results are shown in~\tabref{tab:supple_Abl_adapt}.
We again observe that domain-wise adversaries show strong effectiveness compared to the traditional UDA approaches, confirming that explicitly leveraging the multi-mode nature of target data is essential.
The tendency holds regardless of the UDA methods.
We note that UDA methods in the night domain are even lower than the baseline, which can be interpreted as biased-alignment, as mentioned above.
In contrast, the proposed method outperforms the baseline in every domain, achieving the best-averaged score.
% We confirm the efficacy of the domain-wise adaptation(\ie, Adapt step) with different number of latent target domains($K$). The results with $K=4$ are summarized in the . 
% Similar to results in the main paper, proposed domain-wise adaptation consistently shows better performance over baseline that adopt traditional UDA methods, no matter which UDA methods are utilized. Specifically, night domain adaptation result of this baseline show even worse performance than the model supervised trained on translated source images. On the other hand, domain-wise adaptation not only recover or
% surpass the baseline on night domain, but also shows superior results on final averaged score(\ie, Avg. C and C+O in ~\tabref{tab:supple_Abl_adapt}).
% \clearpage

\subsection{Analysis of the hyperparameter K Selection}
 \label{sec:analysis_K}

\begin{wrapfigure}{R}{0.3\textwidth}
\centering
\vspace{-8mm}
\includegraphics[width=0.28\textwidth]{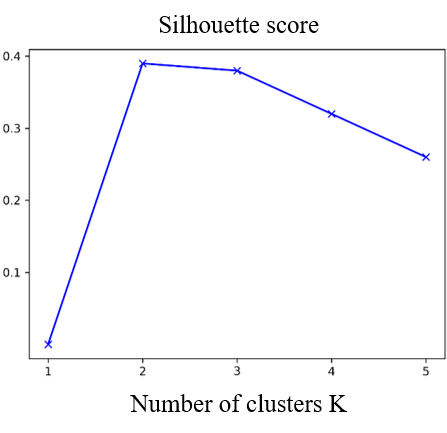}
\vspace{-3mm}
\caption{\label{fig:sil} Silhouette score.}
% \vspace{-6mm}
\end{wrapfigure}

If K value is much less than the optimal, the target distribution might be oversimplified, and some latent domains could be ignored. On the other hand, the images of similar styles might be divided into different clusters, and also each cluster may contain only a few images. In this work, we have set the value of K empirically. Instead, we see one can set the value using existing cluster evaluation metrics such as silhouette score~\cite{rousseeuw1987silhouettes}. It evaluates the resulting clusters by considering the intra-cluster variation and inter-cluster distance at the same time.
As shown in the~\figref{fig:sil}, K=2 and 3 are the strong candidates, and the quality of clusters drops after K=3.

\subsection{Additional Qualitative Results}
 \label{sec:more_qual}

\begin{figure*}%[h*]
\caption{\textbf{Qualitative results.} We provide the semantic segmentation results on the compound domain (``rainy'', ``snowy'', ``cloudy'') and open domain (``overcast''). We can observe clear improvement against both source only and traditional adaptation model~\cite{adaptseg}.}
\centering
\includegraphics[width=0.95\textwidth]{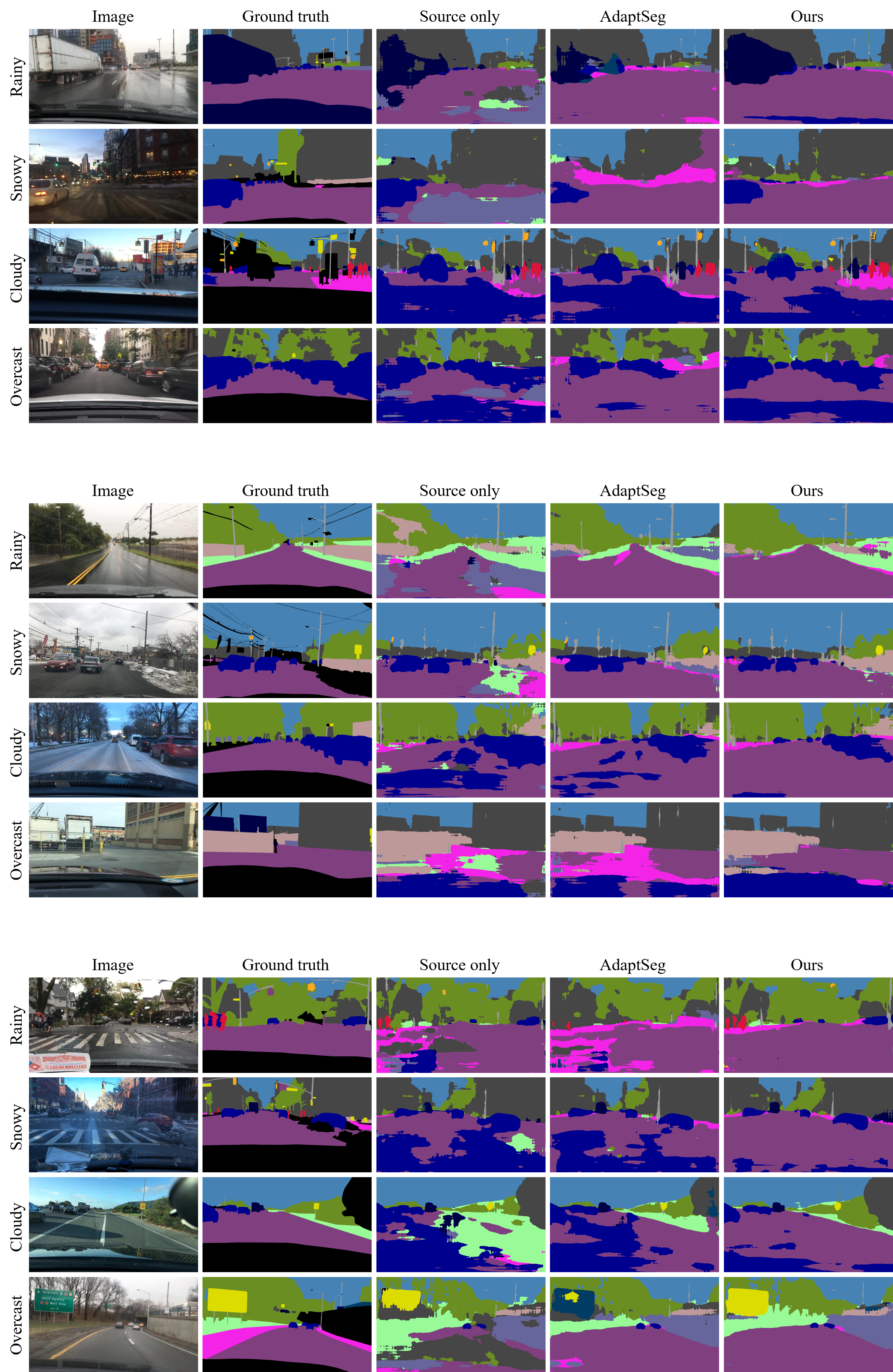}
\label{fig:supple_qual}
% \vspace{-5mm}
\end{figure*}

In~\figref{fig:supple_qual}, we provide more qualitative results.

\subsection{Implementation Details}
 \label{sec:imple_detail}
 
Our model is implemented using Pytorch v0.4.1, CUDNN v7.6.5, CUDA v9.0.

\textbf{Discover step}
We use ImageNet~\cite{deng2009imagenet} pretrained Vgg-16~\cite{simonyan2014very} to encode \textit{style} of target images. Specificallly, we use relu1\_2 features. All target images are resized to have width of 512 pixels while keeping the aspect ratio (\ie, 512$\times$288).
%We use $relu1_2$ features We reduce the dimension of convolutional feature statistics (\ie, mean and standart deviations) to 256 via principal component analysis~\cite{wold1987principal}.

\textbf{Hallucination step}
We detail the two objective functions, $L_{GAN}$ and $L_{sem}$, which are omitted in the main paper.
%We first specify omitted part of hallucination step. 

First, the $L_{GAN}$~\cite{goodfellow2014generative} is defined as follows:

\begin{equation}
\begin{split}
L_{GAN}^{j}(G,D_{I})=  \E_{\mathbf{x}_{S} \sim X_{S}, \mathbf{x}_{T,j} \sim X_{T,j}} logD_{I}(G(\mathbf{x}_{S},\mathbf{x}_{T,j}))  + \E_{\mathbf{x}_{T,j} \sim X_{T,j}} log\left [1 -D_{I}(\mathbf{x}_{T,j}) \right ]
\end{split}
\label{eqn:img_gan_loss}
\end{equation}

Image discriminator $D_{I}$ learns to classify translated source and target images while the generator G tries to produce translated images that are visually similar to target images.
%We using standard GAN loss~\cite{goodfellow2014generative}.

% In order to translate in label-preserved manner, TGCF-DA~\cite{} adopt strong semantic constraint loss $L_{sem}$~\cite{}. 
Second, to enforce strong semantic constraint, the $L_{sem}$~\cite{cycada} is adopted in TGCF-DA~\cite{choi2019self} framework. It is defined as follows:

\begin{equation}
\begin{split}
L_{sem}^{j}(G,f_{seg})=-\E_{(\mathbf{x}_{S},\mathbf{y}_{S})\sim (X_{S},Y_{S}), \mathbf{x}_{T,j} \sim X_{T,j}}\sum_{h,w}\sum_{c}\mathbf{y}_{s}^{(h,w,c)}log(f_{seg}(G(\mathbf{x}_{S},\mathbf{x}_{T,j}))^{(h,w,c)}))
\end{split}
\label{eqn:sem_loss}
\end{equation}

where $f_{seg}$ indicates the semantic segmentation model, which is pretrained on the labeled source domain. Weights of $f_{seg}$ are fixed during training. The loss function strongly encourages the model to preserve the semantics between the source image and the translated image.
%consistency between the images before and after the translation.

In the hallucination step, the source and the target images are resized to 1280$\times$720. For the memory-efficient training, we randomly crop the patches with a resolution of 1024$\times$512. For the testing, we use the original size of 1280$\times$720.

\textbf{Adapt step}
We use segmentation model DeepLab V2~\cite{chen2017deeplab} (for the GTA5/Synscapes experiments) and FCN-8s~\cite{long2015fully} (for SYNTHIA experiments). As noted in the main paper, we use the VGG-16 backbone network. 
For the training, we resize the images of GTA5, Synscapes, and SYNTHIA to 1280$\times$720, 1280$\times$640, 1280$\times$760, respectively~\cite{adaptseg,advent,compound}. We resize the target images in BDD100K to 960$\times$540, following~\cite{compound}.

\begin{table}[h]
\setlength{\tabcolsep}{1pt}
 \centering
 \caption{\textbf{Comparison with the state-of-the-art UDA methods.} We evaluate the semantic segmentation results, Synscapes~\cite{wrenninge2018synscapes} and SYNTHIA~\cite{RosCVPR16} to C-driving~\cite{compound}. For SYNTHIA, we report averaged performance on 16 class subsets following the evaluation protocol used in ~\cite{advent, cbst}.}
    \subfloat[\scriptsize Synscapes to C-driving]
         {
        \resizebox{0.48\textwidth}{!}
        {
        \def\arraystretch{1.1}
        \begin{tabular}{c|ccc|c|cc}
        \hline
        \multicolumn{1}{c}{Source} & \multicolumn{3}{c}{Compound(C)} & \multicolumn{1}{c}{Open(O)} & \multicolumn{2}{c}{Avg.} \\
        Synscapes & Rainy & Snowy & Cloudy & Overcast & C  & C+O \\
        \hline
        \hline
        Source Only & 22.8  & 24.6 & 29.0 & 29.5 & 25.9 & 26.5  \\
        \hline
        CBST~\cite{cbst} & 23.1 & 25.1 & 30.1 & 30.0 & 26.5 & 27.0 \\
        CRST~\cite{zou2019confidence} & 23.1 & 25.1 & 30.1 & 30.1 & 26.6 & 27.1 \\
        AdaptSeg~\cite{adaptseg} & 24.2 & 26.2 & 31.6 & 31.2 & 27.9 & 28.3  \\
        Advent~\cite{advent} & 24.6 & 26.8 & 30.9 & 31.0 & 28.0 & 28.3 \\
        \hline
        Ours & 25.1 & 27.6 & 33.2 & 32.6 & \textbf{29.2} & \textbf{29.6} \\
        \hline
        \end{tabular}
        }
         
         }
    \subfloat[\scriptsize SYNTHIA to C-driving]
         {
        \resizebox{0.48\textwidth}{!}
        {
        \def\arraystretch{1.1}
        \begin{tabular}{c|ccc|c|cc}
        \hline
        \multicolumn{1}{c}{Source} & \multicolumn{3}{c}{Compound(C)} & \multicolumn{1}{c}{Open(O)} & \multicolumn{2}{c}{Avg.} \\
        Synscapes & Rainy & Snowy & Cloudy & Overcast & C  & C+O \\
        \hline
        \hline
        Source Only & 16.3 & 18.8 & 19.4 & 19.5 & 18.4 & 18.5  \\
        \hline
        CBST~\cite{cbst} & 16.2 & 19.6 & 20.1 & 20.3 & 18.9 & 19.1   \\
        CRST~\cite{zou2019confidence} & 16.3 & 19.9 & 20.3 & 20.5 & 19.1 & 19.3\\
        AdaptSeg~\cite{adaptseg} & 17.0 & 20.5 & 21.6 & 21.6 & 20.0 & 20.2  \\
        Advent~\cite{advent} & 17.7 & 19.9 & 20.2 & 20.5 & 19.3 & 19.6 \\
        
        \hline
        Ours & 18.8 & 21.2 & 23.6 & 23.6 & ~\textbf{21.5} & ~\textbf{21.8} \\
        \hline
        \end{tabular}
        }
         
         }
    
\captionsetup{font=footnotesize}
\label{table:supple_SOTA_comparison}
\end{table}

\begin{table*}[h]
\caption{\textbf{Ablation Study on the Adapt step.} The number of latent target domains are set to four (\ie, $K=4$). We again confirm the efficacy of the proposed domain-wise adaptation, demonstrating its superior adaptation results over the direct application of UDA methods~\cite{adaptseg,advent} in compound data.}
\begin{center}

\resizebox{0.70\textwidth}{!}
{
\def\arraystretch{1.5}
\begin{tabular}{cc|cccc|c|cc}
\hline
\multicolumn{2}{c}{} & \multicolumn{4}{c}{Compound(C)} & \multicolumn{1}{c}{Open(O)} & \multicolumn{2}{c}{Avg.} \\
Source &Adapt & Rainy & Snowy & Cloudy & Night & Overcast & C  & C+O \\
\hline
\hline
Ours &None  & 25.2 & 26.4 & 32.7  & 12.1 & 33.8 & 29.1 & 29.5 \\
\hline
Ours &Traditional(~\cite{adaptseg})  & 25.4 & 28.3 & 33.5 & 10.8 & 34.7 & 29.7 & 30.5\\
Ours &Traditional(~\cite{advent})  & 25.9 & 27.8 & 34.2 & 10.6 & 34.7 & 30.1 & 30.7 \\
\hline
Ours &Domain-wise(~\cite{adaptseg})  & 24.6 & 28.8 & 35.0 & 12.0 & 35.1 & \textbf{30.7} & \textbf{30.9} \\
Ours &Domain-wise(~\cite{advent})  & 26.7 & 29.9 & 34.8 & 13.5 & 35.8 & \textbf{31.4} & \textbf{31.8} \\
\hline
\end{tabular}
}

\end{center}
\label{tab:supple_Abl_adapt}
\end{table*}

\end{document}